\documentclass[conference]{IEEEtran}
\IEEEoverridecommandlockouts
\usepackage{cite}
\usepackage{amsmath,amssymb,amsfonts}
\usepackage{algorithmic}
\usepackage{graphicx}
\usepackage{textcomp}
\usepackage{xcolor}
\usepackage{balance}

\usepackage[ruled,vlined, linesnumbered]{algorithm2e}
\usepackage{tabularx}
\usepackage{multirow}
\usepackage{booktabs}

\newcolumntype{Y}{>{\centering\arraybackslash}X}

\def\BibTeX{{\rm B\kern-.05em{\sc i\kern-.025em b}\kern-.08em
    T\kern-.1667em\lower.7ex\hbox{E}\kern-.125emX}}

\begin{document}
\title{Unsupervised Domain Adaptation with Histogram-gated Image Translation \\ for Delayered IC Image Analysis}

\author{\IEEEauthorblockN{Yee-Yang Tee, Deruo Cheng, Chye-Soon Chee, Tong Lin, Yiqiong Shi, Bah-Hwee Gwee}
\IEEEauthorblockA{\textit{School of Electrical and Electronic Engineering} \\
\textit{Nanyang Technological University, Singapore}\\
\{yeeyang.tee, deruo.cheng, chyesoon.chee, lintong, yqshi, ebhgwee\}@ntu.edu.sg}
}

\maketitle

\begin{abstract}
Deep learning has achieved great success in the challenging circuit annotation task by employing Convolutional Neural Networks (CNN) for the segmentation of circuit structures.
The deep learning approaches require a large amount of manually annotated training data to achieve a good performance, which could cause a degradation in performance if a deep learning model trained on a given dataset is applied to a different dataset.
This is commonly known as the domain shift problem for circuit annotation, which stems from the possibly large variations in distribution across different image datasets.
The different image datasets could be obtained from different devices or different layers within a single device.
To address the domain shift problem, we propose Histogram-gated Image Translation (HGIT), an unsupervised domain adaptation framework which transforms images from a given source dataset to the domain of a target dataset, and utilize the transformed images for training a segmentation network.
Specifically, our HGIT performs generative adversarial network (GAN)-based image translation and utilizes histogram statistics for data curation.
Experiments were conducted on a single labeled source dataset adapted to three different target datasets (without labels for training) and the segmentation performance was evaluated for each target dataset.
We have demonstrated that our method achieves the best performance compared to the reported domain adaptation techniques, and is also reasonably close to the fully supervised benchmark.
\end{abstract}


\begin{figure}[htp]
    \centering
    \includegraphics[width=0.8\columnwidth]{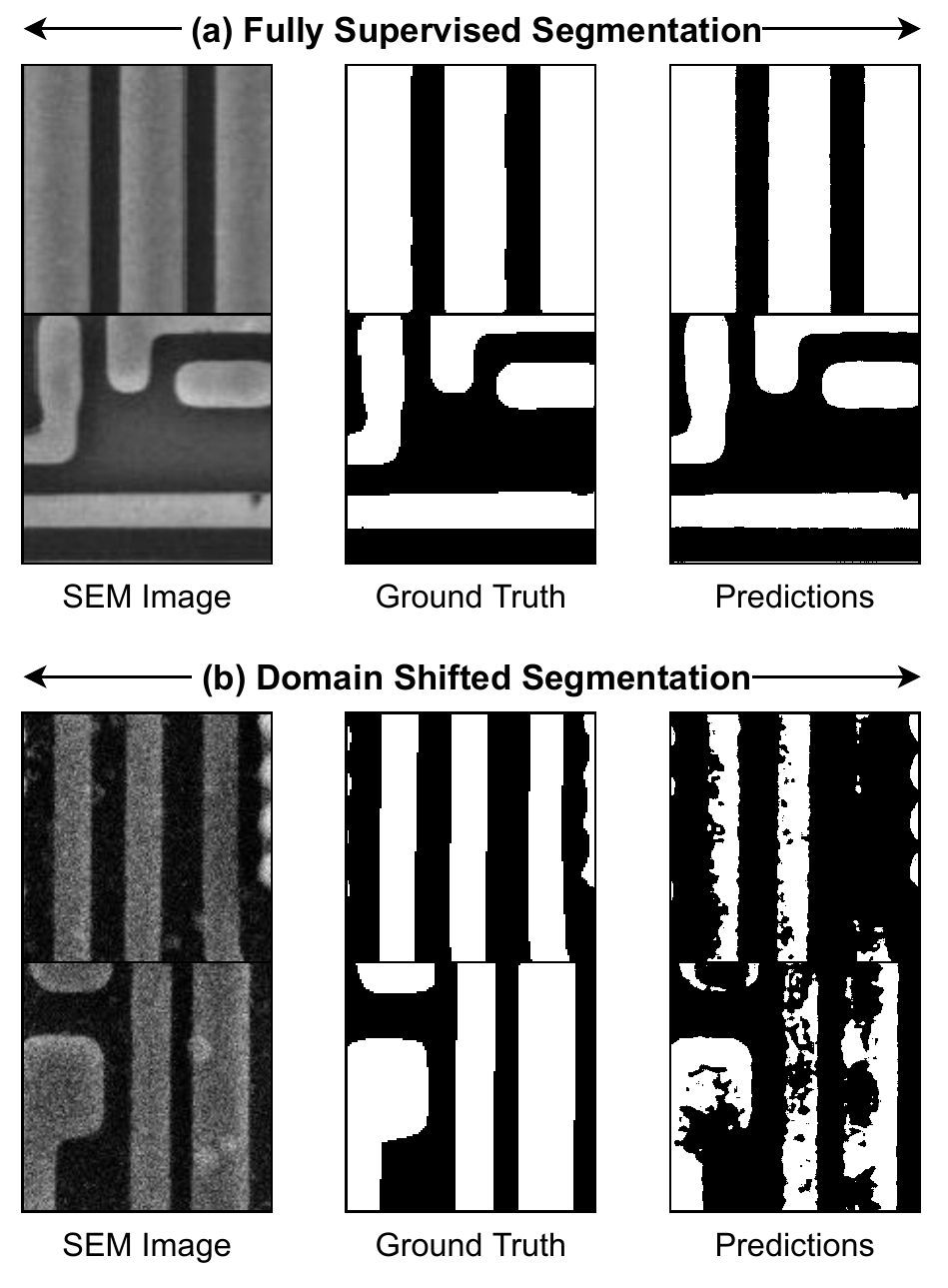}
    \caption{Illustration of the domain shift problem in circuit annotation. (a) A fully supervised segmentation model trained on a given dataset (source) achieves excellent annotation performance on the same dataset. (b) When the same model is applied for the segmentation of another dataset (target), the performance is significantly degraded due to domain shift.}
    \label{fig:naive-transfer-example}
\end{figure}

\section{Introduction}
Integrated Circuit (IC) analysis is a highly reliable approach for hardware assurance, and it is commonly employed for applications such as hardware Trojan detection\cite{zhou2020hardwaretrojanTCAD}, intellectual property protection \cite{knechtel2019intellectual-protection}, and counterfeit detection\cite{ghosh2020counterfeitPAINE}.
Circuit extraction is one of the most accurate methods for manufactured IC analysis, and it consists of several steps, including package removal, chip delayering, imaging, circuit annotation, schematic read-back and final circuit analysis\cite{torrance2009state, botero2021hardware, waite2021preparation, kimura2020silicon, kimura2020decomposition}.
Circuit annotation is a critical step in circuit extraction, as it transforms the image information taken of an actual IC device to a digital representation of all the circuitry in the device.
Fig. \ref{fig:naive-transfer-example}(a) illustrates our circuit annotation task, whereby the scanning electron microscopy (SEM) images (left) are annotated and the objective is to obtain predictions (right) that are as close as possible to the ground truths (middle).

Image processing techniques such as histogram analysis \cite{wilson2020histogramautosegmentation} and machine learning techniques such as support vector machine\cite{cheng2018hybrid} have been reported for the annotation of circuit structures, but their performance could be hampered when encountered with noise, pixel intensity variations, or texture differences in the IC images.
With the rapid rise in popularity of deep learning in recent years, Convolutional Neural Networks (CNN) have been applied for a multitude of tasks in IC and printed circuit board (PCB) analysis \cite{wilt2020deeplearningRFsidechannelPAINE, ghosh2021pcbopticaldeeplearningPAINE,cheng2021tanimoto}.
In \cite{lin2021deep, cheng2021tanimoto, hong2018deep}, deep learning networks such as Fully Convolutional Network (FCN) \cite{long2015fully} and U-Net\cite{ronneberger2015u} were reported for circuit annotation by formulating it as a pixel-wise segmentation task.
The deep learning methods were reported to achieve significantly better performance than the conventional image processing and machine learning methods for circuit annotation. 

However, the performance of deep learning methods could be degraded when applied to circuit images belonging to different ICs or different layers from a single IC.
Fig. \ref{fig:naive-transfer-example}(a) depicts a satisfactory result for circuit annotation that was obtained by a fully supervised model trained on a given circuit dataset.
The performance is far from satisfactory if the same model is utilized for circuit annotation of another IC dataset, as depicted in Fig. \ref{fig:naive-transfer-example}(b). 
The performance degradation is caused by the discrepancy in distribution of the IC images from the different datasets.
This could be due to differences in distribution of pixel intensities, the amount of contrast between background and foreground, the presence of noise, etc.
This is commonly known as the domain shift challenge in circuit images, which could stem from variations in the IC manufacturing process, sample preparation process, or imaging process \cite{botero2021hardware}.
The conventional way to tackle the domain shift challenge is by repeating the fully supervised method, including the data collection and labeling process, for every new dataset to be analyzed.
However, this could be impractical as it requires a large amount of manual effort and has to be repeated for different ICs or for different layers from a single IC.

To address the above-mentioned domain shift problem, we intend to utilize an IC dataset with known labels (denoted as the source dataset) for the training of a segmentation model, and adapt it to another dataset without labels (denoted as the the target dataset).
This is known as the unsupervised domain adaptation scenario.
We explore techniques which can leverage labeled data in the source domain and transfer the learned knowledge to adapt the model to the target domain\cite{pan2009surveyontransferlearning}.
The aim is to eliminate the need for manual labeling in the target dataset when training a deep learning model for circuit annotation, whilst still achieving an acceptable segmentation accuracy.

In this paper, we propose Histogram-gated Image Translation (HGIT), an unsupervised domain adaptation framework for circuit annotation.
Our method leverages generative adversarial networks (GAN)\cite{goodfellow2014generative} for image translation, which are trained on just the unlabeled source and target circuit images, and then utilized to transform the source images into the domain of the target images.
We then perform a statistical comparison between the histograms of the transformed images and the target dataset to curate the transformed images.
The curated transformed images are paired with the source masks for the training of a segmentation model for circuit annotation.
The trained segmentation model is subsequently applied to the target circuit images for circuit annotation.

\begin{figure*}[htp]
    \centering
    \includegraphics[width=1.8\columnwidth]{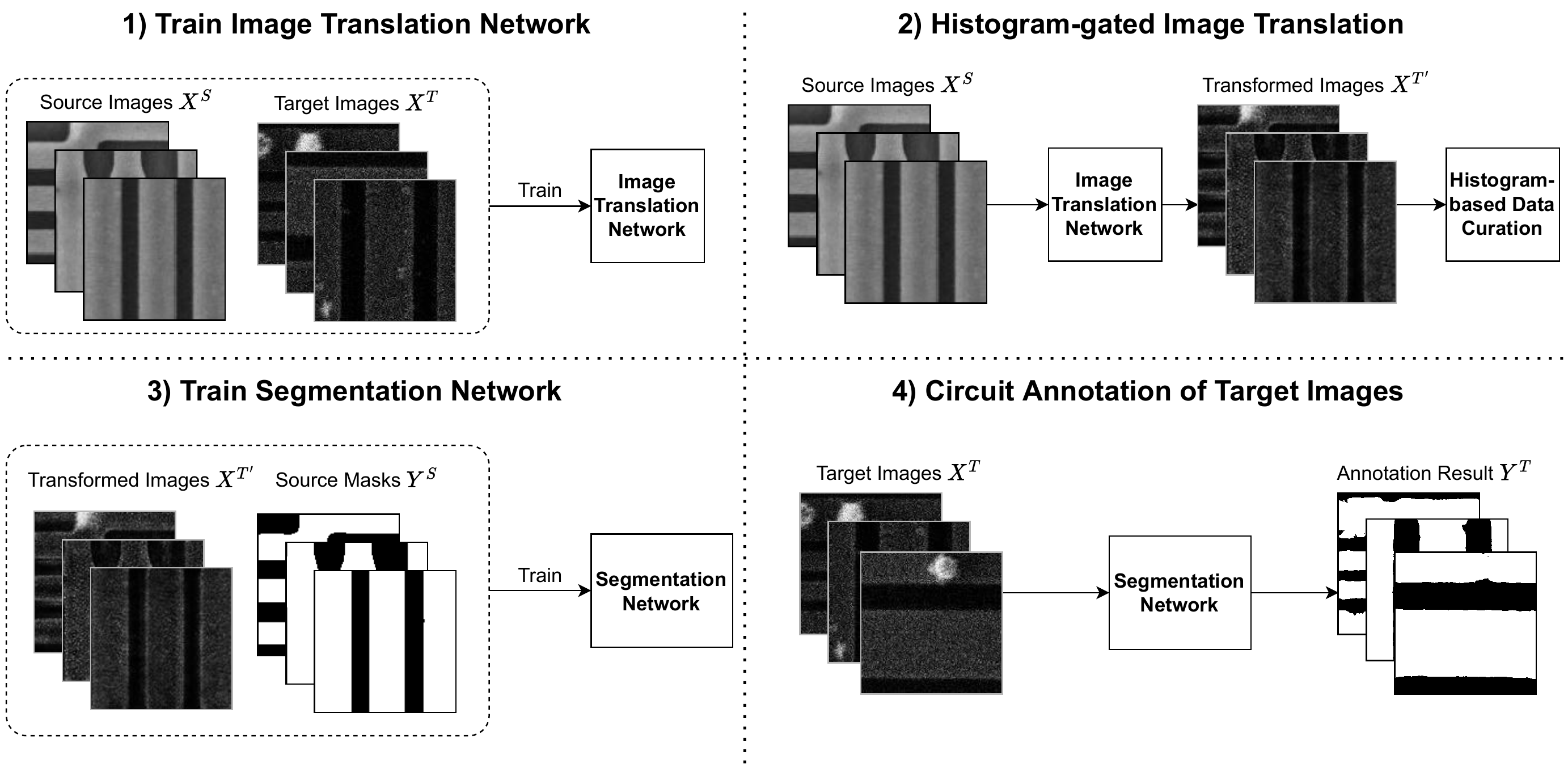}
    \caption{Overview of our unsupervised domain adaptation framework for circuit annotation which leverages GAN-based image translation and histogram-based data curation.}
    \label{fig:method-main}
\end{figure*}

The contributions of this paper are summarized as follow:
\begin{itemize}
    \item We propose HGIT, an unsupervised domain adaptation framework for circuit annotation that combines GAN-based image translation and histogram-based data curation.
    \item Our HGIT demonstrated the best performance as compared to the reported domain adaptation techniques, achieving a 8.76\% improvement in segmentation performance as compared to the second best method.
    \item Our HGIT does not require any additional data labeling on the target dataset, significantly saving time and effort when performing the analysis on new datasets of different ICs or different layers.
    \item To the best of our knowledge, this is the first work utilizing unsupervised domain adaptation for IC image analysis.
\end{itemize}

\section{Related Work}
Unsupervised domain adaptation leverages sufficient labeled data in a source domain and transfers the learned knowledge to adapt a segmentation model to the target domain.
It has attracted significant research interest in recent years due to its ability to remove the need for manual annotation on new datasets.
Domain adaptation involves the alignment of the domain shift between the distributions of a known source dataset and a new or unknown target dataset.
Image translation, feature matching, or a combination of image and feature matching have been reported for unsupervised domain adaptation. 

Image translation is popularly adopted for domain adaptation, which could be achieved with techniques such as image processing and generative adversarial networks (GAN)-based\cite{goodfellow2014generative} methods.
Conventional image processing techniques were adopted for image translation in \cite{ma2020histogram}, where histogram matching was utilized to align the histogram statistics of the source domain with the target domain. 
Histogram matching involves transforming the probability density function of pixel intensities in the source images to become the same as the target images.
However, histogram matching could have a degraded image translation performance in IC images with few circuit structures, whereby enforcing a specific pixel intensity distribution could cause unwanted bright or dark regions in the transformed images, which is not ideal for segmentation model training.

GAN-based techniques have achieved promising performance for image translation in multiple domains \cite{zhu2017cyclegan, kim2020cvpr_textureinvariantdomainadapt,sandfort2019cyclegan_ctsegmentation}.
CycleGAN\cite{zhu2017cyclegan} employs a GAN-based framework with a cycle consistency loss to learn the transformation from source to target domain (and vice-versa) while preserving the semantic content of the images.
The transformation is achieved by training a pair of generator networks with a cycle consistency loss that enforces that the image from the original domain can be recovered after performing a two way translation.
Adversarial training with a discriminator network\cite{goodfellow2014generative} is also implemented in CycleGAN to achieve good realism in the transformed images.
However, due to the stochastic nature of neural network training, it can be difficult to ensure the quality of all the transformed images without performing visual inspection of the transformed results.

Fourier domain adaptation (FDA) proposed in \cite{yang2020FDA} utilized Fourier Transform to transfer the frequency features of the target domain to the source domain to achieve image translation.
The low frequency part of the source amplitude is replaced with the corresponding region from the target amplitude and an inverse Fourier Transform is then computed to obtain the transformed image.
One key advantage of FDA is that it does not require the training of a deep neural network for the translation step. 
However, the transformation quality could be degraded on circuit images that have a lower visual complexity or a small spatial dimension.

The Simplified Unsupervised Image Translation (SUIT) technique proposed in \cite{li2020SUIT} combines feature and image alignment for effective domain adaptation.
A consistency loss achieved through a separate segmentation network and a feature extractor (using pre-trained CNN) was introduced to improve the feature alignment of source and target domain.
The consistency loss is utilized to optimize the generator network to improve the visual appearance of the transformed images.

\section{Methodology}
In unsupervised domain adaptation for circuit annotation, we denote the labeled source domain as $\mathcal{D}^{S} = \{(X^S,Y^S)\}$ and the unlabeled target domain as $\mathcal{D}^{T} = \{X^T\}$, where $X^S$ is the set of all source images, $Y^S$ is the set of all source masks (ground truth labels) associated with $X^S$, and $X^T$ is the set of target images.
The goal is to utilize the labeled source domain and unlabeled target domain to train a segmentation network to annotate the target images, without performing any manual labeling of the target images.
Fig. \ref{fig:method-main} depicts our unsupervised domain adaptation framework for circuit annotation, consisting of 4 main steps including training of image translation network, histogram-gated image translation, training of segmentation network, and circuit annotation of target images.
We now delineate each component of our framework in the following sections.

\subsection{Train Image Translation Network}
We adopt the CycleGAN framework \cite{zhu2017cyclegan} for image translation from the source to target domain.
The goal is to obtain an image translation network that can transform the source images into the style of the target images.
To achieve this, a pair of encoder-decoder networks ($G_{S\rightarrow T}$, $G_{T\rightarrow S}$) are trained to transform images from source to target domain and target to source domain respectively.
Both networks are trained on the source and target images $X^S$ and $X^T$ in an unsupervised manner with the cycle consistency loss $L_{cyc}$ given in

\begin{eqnarray} \label{eqn:cycle-consistency-loss}
  L_{cyc} = \mathbb{E}||X^{S} - G_{T\rightarrow S}(G_{S\rightarrow T}(X^{S}))|| 
  \notag\\+ \mathbb{E}||X^{T} - G_{S\rightarrow T}(G_{T\rightarrow S}(X^{T}))||
\end{eqnarray}
where $\mathbb{E}$ represents the mean value computed across the training samples.

\subsection{Histogram-gated Image Translation}\label{section:method-hgit}
Image translation is performed by $G_{S\rightarrow T}$ to transform the source images into the style of the target images, obtaining the transformed images $X^{T'}$ given by (\ref{eqn:image-translation-inference}). $X^{T'}$ should retain the same semantic content (i.e., metal lines should have the same location) as the source images $X^S$, while having a similar visual appearance to the target domain.

\begin{equation} \label{eqn:image-translation-inference}
   X^{T'} = G_{S\rightarrow T}(X^{S})
\end{equation}

The transformed images that have a smaller domain gap to the actual target images are then selected with histogram-based data curation.
The histogram distribution of all the target images are first computed and averaged to obtain the mean histogram profile.
Next, a Kolmogorov-Smirnov\cite{kolmogorov-smirnov} statistical test was performed on the histogram distribution of each transformed image against the average histogram distribution of the target dataset.
The p-values obtained are used as a similarity measure between the transformed images and target images.
The transformed images are then curated by selecting the top $N\%$ of images for the subsequent training of a segmentation network, and we set the value of $N$ to 70 in our experiments.

\subsection{Train Segmentation Network}
The transformed images $X^{T'}$ are paired with the source masks $Y^S$ to train a segmentation network $F$.
Our unsupervised domain adaptation framework is architecture agnostic and different segmentation network architectures can be used.
For our metal line circuit annotation task, the pixel-wise binary cross entropy loss given in (\ref{eqn:cross-entropy-loss}) is utilized to optimize the network parameters. 

\begin{equation}\label{eqn:cross-entropy-loss}
L_{BCE} = - \mathbb{E} ||\log{(Y^S \cdot F(X^{T'}))}||
\end{equation}
where $F$ represents the segmentation network.

\subsection{Annotation of Target Images}
The trained segmentation network is finally applied for the annotation of the target images $X^T$.
This produces the predicted target masks which are the annotation results $Y^T$, as given in (\ref{eqn:annotation-inference}).
Error correction can be performed on the predicted target masks, and the subsequent steps of circuit extraction can be carried out.

\begin{equation} \label{eqn:annotation-inference}
   Y^{T} = F(X^{T})
\end{equation}

\section{Experiments}
We have evaluated our proposed framework in comparison with several reported techniques for domain adaptation.
In another setting, we also establish an upper bound benchmark through a fully supervised method for additional comparison.
The pixel-wise segmentation accuracy (SA) and Intersection-over-Union (IoU) was evaluated and utilized as the comparison metric across models.
In the following sections, we describe the dataset and implementation details, and then present the experimental results.

\begin{table}[htp]
  \centering
  \caption{Summary of the four datasets utilized in our experiments.} 
 \label{table:dataset-description}

\bgroup
\def\arraystretch{1.5}
\begin{tabularx}{\columnwidth}{YYYYYY}
\toprule
Device Name & Type of Device& Layer Name & Training Images & Testing Images & Domain Type\\
\midrule
Device-1 & Micro-controller & Metal 2 & 1082 & - & Source \\
Device-1 & Micro-controller & Metal 1 & 1098 & 274 & Target\\ 
Device-2 \phantom{asdadas} & FPGA & Metal 6 & 1,152 & 288 & Target\\ 
Device-3 \phantom{asdadas} & FPGA & Metal 3 & 1,229 & 307 & Target\\
\bottomrule

\end{tabularx}
\egroup
\end{table}

\subsection{Dataset}
In our experiments, we utilize 4 datasets from different ICs or different layers, as summarized in Table \ref{table:dataset-description}.
For each dataset, there are around 1,500 total images of dimension 128$\times$128 pixels, and we utilize a train test split of 80:20.
The Device-1 Metal 2 (M2) layer is designated as the source dataset while the 3 other datasets from different ICs or different layers are designated as the target datasets.
The pixel-wise ground truth labels were prepared using the semi-automated labeling method from \cite{lin2021deep}.

\begin{table*}[htp]
  \centering
  \caption{Segmentation results for three target datasets of different devices, with training data that was adapted from a single source dataset (Device-1 M2).} 
 \label{table:results-main}

\bgroup
\def\arraystretch{1.4}%
\begin{tabularx}{2\columnwidth}{cYYYYYYYY}
\toprule
Target Dataset  & \multicolumn{2}{c}{Device-1 M1} & \multicolumn{2}{c}{Device-2 M6} &\multicolumn{2}{c}{Device-3 M3} &\multicolumn{2}{c}{Averaged} \\
\midrule[\heavyrulewidth]

Method  & SA & IoU \(^{\mathrm{\dagger}}\) & SA & IoU & SA & IoU & SA & IoU  \\
\cmidrule(lr){1-1} \cmidrule(lr){2-3} \cmidrule(lr){4-5} \cmidrule(lr){6-7} \cmidrule(lr){8-9}

Source Only & 0.9025 & 0.8275 & 0.3960 & 0.0877 & 0.8473 & 0.7812 & 0.7153 & 0.5655 \\
Histogram Matching \cite{ma2020histogram} & 0.9131 & 0.8637 & 0.6151 & 0.5063 & 0.8055 & 0.7528 & 0.7779 & 0.7076 \\
FDA \cite{yang2020FDA} & 0.9688 & 0.9341 & 0.6454 & 0.5577 & 0.8525 & 0.8028 & 0.8222 & 0.7648 \\
SUIT \cite{li2020SUIT}& 0.8380 & 0.7207 & 0.6718 & 0.5658 & 0.7589 & 0.6559 & 0.7562 & 0.6475 \\
CycleGAN \cite{zhu2017cyclegan}& 0.9793 & 0.956 & 0.7705 & 0.6902 & 0.7821 & 0.6867 & 0.8440 & 0.7776 \\
\textbf{HGIT (Ours)}  & \textbf{0.9867} & \textbf{0.9646} & \textbf{0.8066} & \textbf{0.7124} & \textbf{0.9512} & \textbf{0.9187} & \textbf{0.9148} & \textbf{0.8652} \\
\midrule
Fully Supervised (Upper Bound) & 0.9935 & 0.9825 & 0.9747 & 0.9564 & 0.9758 & 0.9543 & 0.9813 & 0.9644 \\

\bottomrule

\multicolumn{9}{l}{\hspace*{0pt}\(^{\mathrm{\dagger}}\) SA: Pixel-wise segmentation accuracy, IoU: Intersection-over-Union. Metrics were averaged over 5 runs initialized with random seeds.}

\end{tabularx}
\egroup
\end{table*}

\begin{figure*}[htp]
    \centering
    \includegraphics[width=1.8\columnwidth]{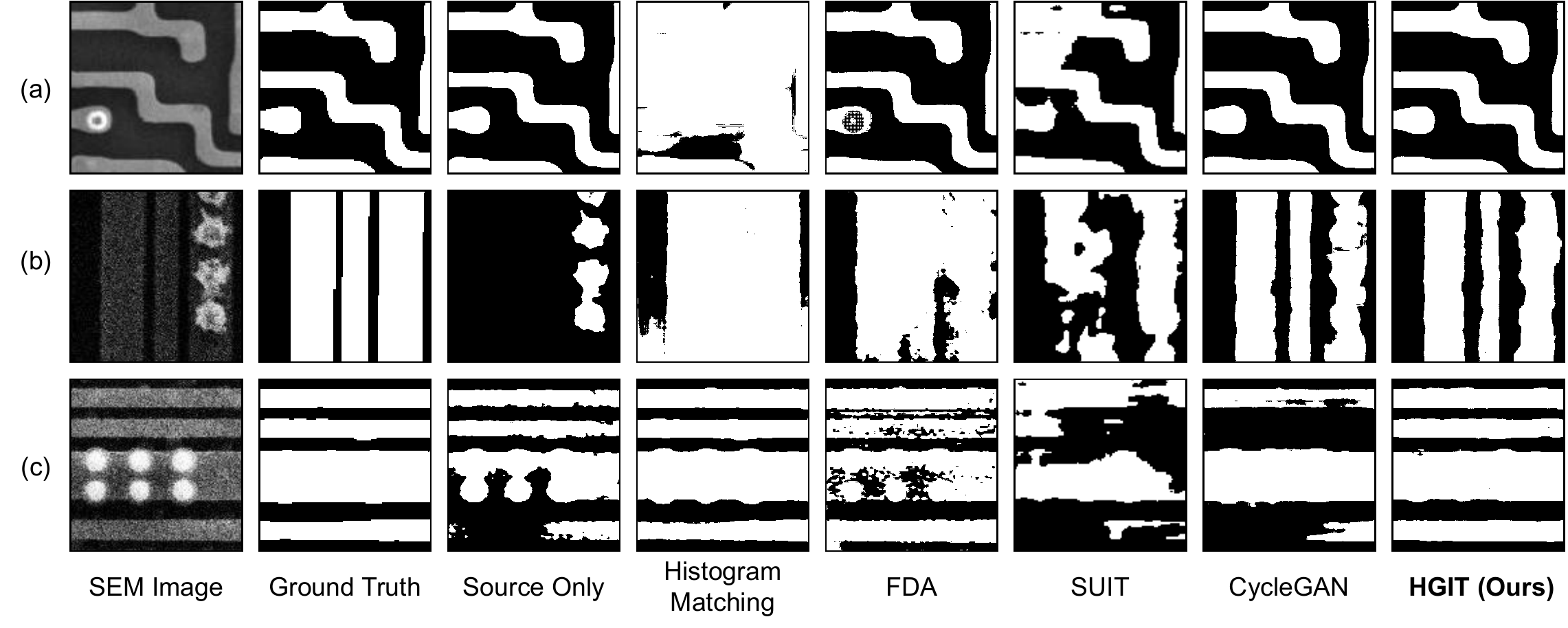}
    \caption{Visualized segmentation results obtained by each reported technique and our proposed HGIT when applied to a sample from each target dataset: (a) Device-1 M1, (b) Device-2 M6, (c) Device-3 M3.}
    \label{fig:prediction-visual-results}
\end{figure*}

There are 3 distinct evaluation scenarios in our experiments, including source only, unsupervised domain adaptation, and fully supervised.
The source only scenario involves the training of a segmentation network only on the source training images and masks (ground truth labels), and this is commonly used as a baseline in unsupervised domain adaptation works \cite{li2020SUIT}.
The trained segmentation network is then applied to the 3 different sets of target testing images for evaluation.
The unsupervised domain adaptation scenario includes the 4 reported techniques (histogram matching\cite{ma2020histogram}, FDA\cite{yang2020FDA}, SUIT\cite{li2020SUIT}, and CycleGAN\cite{zhu2017cyclegan}) and our proposed HGIT.
For the unsupervised domain adaptation scenario, the source training images are adapted to the target domain, and then paired with the source training masks for training a segmentation network.
The trained segmentation network is then applied to the respective set of target testing images for evaluation, and this is repeated for the 3 different target datasets.
Lastly, the fully supervised scenario is implemented as an upper bound benchmark. 
In this scenario, a segmentation network is trained on the target training images and masks, and the trained segmentation network is applied to the respective set of target testing images for evaluation.

\subsection{Implementation Details}\label{section:implementation}
All experiments in this paper were implemented with the PyTorch\cite{pytorch} framework on a single NVIDIA A5000 GPU\footnote{With this particular computational hardware, the average inference time for a single image (128$\times$128) from our dataset was 0.00689 seconds.}.
We now describe the implementation details for each of the reported domain adaptation techniques.
For histogram matching, the histogram statistics of the source training images were transformed to the average histogram profile of the target training images.
For FDA, we utilized the publicly available implementation\footnote{Available at https://github.com/YanchaoYang/FDA} and transform each of the source images to a randomly sampled target image to be used for training.
For CycleGAN, a transformation network was trained on the source training images and target training images using the publicly available implementation\footnote{Available at https://github.com/junyanz/pytorch-CycleGAN-and-pix2pix}.
The source training images were then translated with the transformation network into the style of the target domain.
The transformed images (in the style of the target domain) were then utilized to train a segmentation network with the source masks as the ground truth labels.
For the SUIT technique, training of a transformation and segmentation network was performed in a single stage using the source training images, source masks and target training images as inputs.
The trained segmentation network was utilized for evaluation on the target testing dataset.
For our proposed HGIT, image translation was performed on the source training images, and the top 70\% of transformed images were selected for further training based on the statistical similarity of histogram distribution to the average target histogram.
The selected transformed images were then utilized to train a segmentation network with the source mask as ground truth labels.

\begin{figure*}[htp]
    \centering
    \includegraphics[width=1.5\columnwidth]{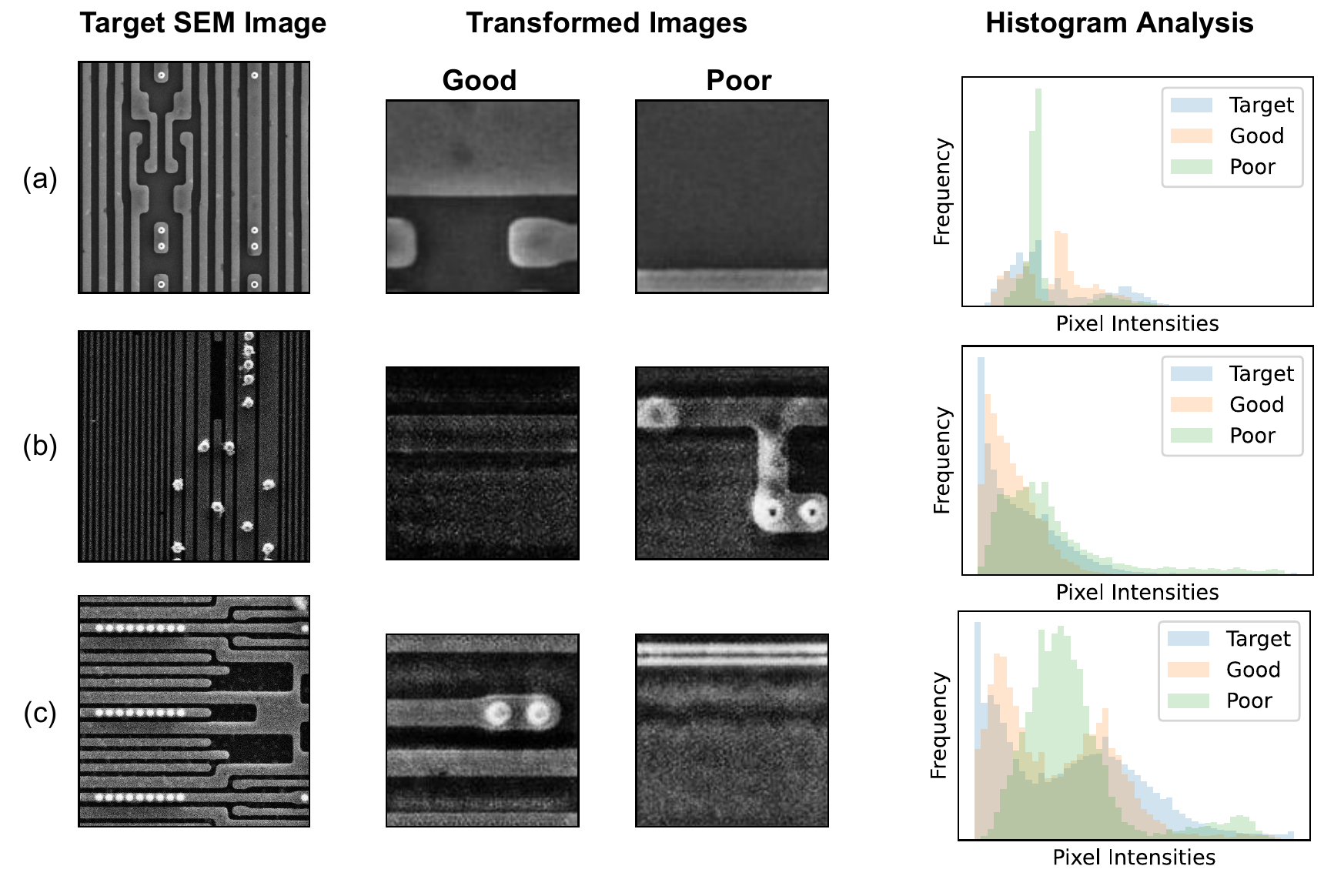}
    \caption{A zoomed out sample of each target domain is shown on the leftmost column. The image translation results are shown as a 'Good' transformed image and 'Poor' transformed image in the middle column. The histograms of the average target domain images, and the two transformed images are plotted on the right. Each row represents a separate target domain: (a) Device-1-M1, (b) Device-2-M6, (c) Device-3-M3. }
    \label{fig:domain-gap-filtering}
\end{figure*}

\subsection{Evaluation Metrics}
We utilize two popular metrics for evaluation and comparison in our experiments, namely the pixel-wise segmentation accuracy (SA) and Intersection-over-Union (IoU) \cite{wang2020imagesegmentationmetrics}.
Circuit annotation is a single class, pixel-wise classification of metal lines.
We first define several key terms:
\begin{itemize}
    \item True Positive (TP): Number of pixels correctly classified as metal line 
    \item False Positive (FP): Number of incorrectly classified as metal line
    \item True Negative (TN): Number of pixels correctly classified as not metal line
    \item False Negative (FN): Number of pixels incorrectly classified as not metal line
\end{itemize}
Following that, the SA and IoU metrics are given in (\ref{eqn:segmentation-accuracy-metric}) and (\ref{eqn:iou-metric}) respectively

\begin{equation}\label{eqn:segmentation-accuracy-metric}
SA = \frac{TP + TN}{TP + FP + TN + FN}
\end{equation}

\begin{equation}\label{eqn:iou-metric}
IoU = \frac{TP}{TP + FP + FN}
\end{equation}

\subsection{Comparison with Reported Domain Adaptation Techniques}\label{section:mainresult}
Table \ref{table:results-main} presents the segmentation results in terms of the pixel-wise segmentation accuracy (SA) and Intersection-over-Union (IoU), across 3 different target datasets.
The averaged metrics across the 3 datasets are presented for an overall comparison.
Our proposed HGIT has achieved the highest performance in comparison to all the reported techniques.
When averaged across the 3 datasets, we have achieved an improvement of 7.08\% and 8.76\% in the pixel-wise segmentation accuracy and IoU respectively, as compared to the second best reported technique (CycleGAN).
Furthermore, our method has achieved a reasonably close performance to the upper bound benchmark for the Device-1 M1 set (less than 2\% difference) and the Device-3 M3 set (less than 4\% difference).
For the Device-2 M6 dataset, all the reported techniques have achieved a significant improvement as compared to the baseline method (Source Only), but the absolute segmentation performance is far from the upper bound benchmark.
This is because the Device-2 M6 images have a much lower pixel intensity (darker in appearance) and a lower contrast between circuit structures as compared to the source dataset, which results in a larger domain gap between the two datasets.

Fig. \ref{fig:prediction-visual-results} depicts the segmentation results of the reported domain adaptation methods and our proposed technique when applied to a sample from each of the 3 target domains.
From Fig. \ref{fig:prediction-visual-results}, we can see that that our HGIT could perform the metal line annotations on each of the sample target images without incorrect breakage or joining of metal lines, and has achieved the best segmentation performance as compared to the other reported techniques.

\subsection{Study on Histogram-gating and Image  Translation}\label{section:image-translation-study}
The objective of image translation is to transform the images from the source domain into a transformed image that is similar to the target domain.
A zoomed out sample of each target domain is shown on the left of Fig. \ref{fig:domain-gap-filtering}.
Our proposed HGIT performs image translation with CycleGAN on the source images to the target domain, and two examples of such transformed images are depicted in the middle of Fig. \ref{fig:domain-gap-filtering}.
Out of the two transformed images, the 'Good' column is visually similar to the target image, and also contains some circuit structures such as metal lines for segmentation model training.
However, CycleGAN could produce images that are not optimal for training, as depicted in the 'Poor' column in Fig. \ref{fig:domain-gap-filtering}.

Image translation with GANs generally produce realistic transformed images that are highly similar to the target domain, but there could still be failure cases.
This could be a result of insufficient circuit structures as seen in the poor transformed images in Fig. \ref{fig:domain-gap-filtering} for target domain (a) and (c) where the majority of the transformed image is the background or metal line only.
Another failure case can be seen in Fig. \ref{fig:domain-gap-filtering}(b) whereby the metal line regions in the transformed images were abnormally brightened, which could be due to the stochastic nature of GAN training. 
This demonstrates the importance of curating the transformed images before utilizing them for training a segmentation network in an unsupervised domain adaptation framework.

The histogram statistics of the averaged target domain are also presented, along with the histogram statistics of the two transformed images.
We can observe that the poorly transformed images have a significant difference in the histogram distribution as compared to the good transformed images.
In the histogram-based data curation step (See earlier Section \ref{section:method-hgit} for more details), we curate the optimal transformed images for training a segmentation network, and the poor transformed images exemplified in Fig. \ref{fig:domain-gap-filtering} were successfully removed by our proposed HGIT.

\section{Conclusion and Future Work}
In this paper, we have proposed HGIT, an unsupervised domain adaptation framework for delayered IC image analysis.
We have demonstrated its effectiveness as a broad method that can avoid labor intensive manual data labeling in circuit annotation.
HGIT performs GAN-based image translation to transform the source images to be similar to the target images, and we curate the good translation results for further training based on histogram statistics.
We have evaluated our proposed HGIT in comparison with four reported domain adaptation techniques on three different target datasets obtained from different ICs.
Our method has achieved the best performance out of all the reported techniques, and it is also reasonably close to the fully supervised benchmark.
Our HGIT and unsupervised domain adaptation methods in general are promising for greatly reducing the amount of manual effort in practical applications of delayered IC image analysis. 

In a future work, it would be valuable to evaluate a larger number of IC datasets with different image distributions, and study the interactions between different types of source and target datasets.

\section*{Acknowledgment}
We thank Microelectronics Failure Analysis (MFA) Lab, School of Materials Science and Engineering (MSE), Nanyang Technological University, for providing the microscopic images of our targeted devices.

\bibliographystyle{IEEEtran}\
\balance
\bibliography{ref}

\end{document}